\documentclass[runningheads]{llncs}
\usepackage{graphicx}
\usepackage[utf8]{inputenc}
\usepackage[T5]{fontenc}
\usepackage{amsmath}
\usepackage{hyperref}
\usepackage{graphicx}
\usepackage{amsmath}
\usepackage{amssymb}
\usepackage{multirow}
\usepackage{tabularx}
\usepackage{blindtext}
\usepackage{array}
\usepackage{graphicx}
\usepackage[misc]{ifsym}
\usepackage{amsmath}

\begin{document}

\title{Multi-stage Information Retrieval for Vietnamese Legal Texts}


\author{Nhat-Minh Pham\inst{1,2}\and Ha-Thanh Nguyen\inst{3} \and Trong-Hop Do \inst{1,2}}
\authorrunning{Nhat-Minh Pham et al.}
%
\institute{University of Information Technology,Ho Chi Minh City, Vietnam \and Vietnam National University, Ho Chi Minh City, Vietnam \and National Institute of Informatics, Tokyo, Japan \\
\email{18520102@gm.uit.edu.vn, nguyenhathanh@nii.ac.jp, hopdt@uit.edu.vn}\\}

\maketitle              
\begin{abstract}
This study deals with the problem of information retrieval (IR) for Vietnamese legal texts. Despite being well researched in many languages, information retrieval has still not received much attention from the Vietnamese research community. This is especially true for the case of legal documents, which are hard to process. This study proposes a new approach for information retrieval for Vietnamese legal documents using sentence-transformer. Besides, various experiments are conducted to make comparisons between different transformer models, ranking scores, syllable-level, and word-level training. The experiment results show that the proposed model outperforms models used in current research on information retrieval for Vietnamese documents.
\end{abstract}
\section{Introduction}
Legal information retrieval is a specialized task in natural language processing (NLP) that involve retrieving relevant legal documents \footnote{In this paper, "text" and "document" are used interchangeably}  given a query. Compared with traditional text retrieval, legal text retrieval is more difficult since legal documents are often long and complicated. On the other hand, questions are often highly complex and need a person with good expertise to give the correct answer.

Legal information retrieval for English documents has received a lot of attentions. COLIEE \footnote{\url{https://sites.ualberta.ca/~rabelo/COLIEE2022/}} is an annual competition about legal information extraction/entailment or JURIX \footnote{\url{http://jurix.nl/conferences/}}  is an annual conference on legal AI.
However, there are very few studies on Vietnamese legal information retrieval. Recent studies of Vietnamese legal text retrieval have been mainly on CNN architecture combined with attention mechanism \cite{kien-etal-2020-answering,nguyen2022toward}, using fine-tuned Multilingual transformer model like XLM-RoBERTa \cite{nguyen2022toward}. In recent years, monolingual transformer models like PhoBERT \cite{nguyen-tuan-nguyen-2020-phobert}, ViBERT \cite{tran2020improving} have beat the state-of-the-art (SOTA) in many NLP tasks in Vietnamese like Part-of-Speech(POS tagging), Named-Entity Recognition(NER), and Dependency Parsing(DP). Sentence-BERT(SBERT) \cite{reimers2019sentence} and models evolved from it \cite{thakur-etal-2021-augmented} have recently reached significant results on several tasks like semantic searching or information retrieval. In this paper, we focus on exploring fine-tuned sentence-transformer monolingual models for Vietnamese legal text retrieval.

The contribution of this paper is two-fold. First, we propose a novel pipeline of multi-stage information retrieval based on sentence-transformer for Vietnamese legal texts. Experiments were conducted and the results show that the proposed pipeline significantly outperforms existing models. Second, empirical analyses of the effects of multiple factors including language models(LM), word segmentation methods, and ranking scores on the performance of information retrieval are performed. The results show that SPhoBERT-large language model, word-based segmentation, and system with combine ranking scores yield the best result in information retrieval for Vietnamese legal documents. 

\section{Background and Related Work}
In this paper, we focus on answering legal queries at the article level. Given a legal query, our goal is to retrieve all the relevant articles, that can be used as the fundamental to answer the query.

Many approaches, especially for ad-hoc text retrieval have been proposed, from past decades to recent years. 

\subsubsection{Non-neural approaches} These methods, decide relevance based on the frequency and occurrence of the words in the query and the documents. Non-neural approaches do not perform well in the semantic searching problem due to lexical mismatching in the answers. However, they are still useful for current SOTA models. BM25 \cite{10.1145/1031171.1031181} and tf-idf \cite{SALTON1988513} are well-known among all while BM25+\cite{robertson1994some} is currently the most effective in the approach.  To overcome the lexical mismatching challenge, dense embedding is used to represent queries and documents. The main idea of this method was proposed with  the LSI approach \cite{Deerwester1990IndexingBL}. 
\subsubsection{Attention mechanism approaches} Attention mechanism  makes the model able to focus more on the main keywords or informative sentences in the original text. Kien and Nguyen et al. signed a simple attentive convolution neural network for Vietnamese legal text retrieval {\cite{kien-etal-2020-answering}}. Nguyen also used this method in his dissertation{\cite{nguyen2022toward}}. Their retrieval system can capture both local and global contexts to construct to build representation vectors. 

\subsubsection{Transformer cross-encoder approaches} BERT \cite{devlin2018bert} brings breakthroughs in NLP. The cross-encoder approach is the first widely approach of BERT in information retrieval. Both query and document will be passed simultaneously to the BERT network. Birch \cite{akkalyoncu-yilmaz-etal-2019-applying} - the system combines lexical matching and the cross-encoder approach for better performance. Several other studies \cite{lee-etal-2019-latent,karpukhin-etal-2020-dense,laskar-etal-2020-contextualized} applied the cross-encoder approach and reached significant results. BERT-PLI \cite{shao2020bert} is a retrieval system based on these approaches. However, these approaches require lots of time for training and costly computation resource.
\subsubsection{Transformer bi-encoder approaches} Cross-encoder approaches are costly and time-consuming for training. Motivated by this, Reimers and Gurevych presented SBERT which uses SiameseBERT-network to represent semantically meaningful sentence embeddings. SBERT produce vector embedding for each sentence independently. When we want to compare two sentences, we just need to calculate the cosine similarity of the two existing vectors. The authors of SBERT train the bi-encoder from BERT and  compare two input sentences using cosine similarity. For question answering(QA) or IR tasks, many studies developed from or applied SBERT \cite{SBERT-WK,condor2021automatic} and get high performance.According to recent research by Gao and Callan \cite{gao-callan-2021-Condenser} standard LMs' internal attention structure is not ready to use for dense encoders. The authors proposed a new transformer architecture, Condenser as an improvement for the bi-encoder approach.  For applying bi-encoder approach for Vietnamese, to the best of our knowledge, there is only one paper \cite{ha2021utilizing} on bi-encoder approach and the solution \footnote{\url{https://github.com/CuongNN218/zalo_ltr_2021}} for an annual competition of artificial intelligence. 

\section{Sentence-transformers based Multi-stage information retrieval for Vietnamese legal text}
\begin{figure}
    \raggedright 
    \centering
    \includegraphics[scale=0.6]{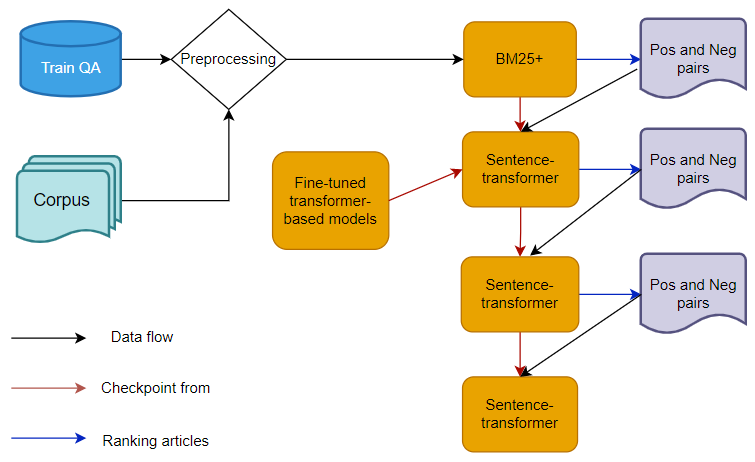}
    \caption{Our proposed pipeline for training}
    \label{pipeline for training}
\end{figure}

The general idea of our approach is to use both lexical matching and semantic searching to improve the performance of our system. For lexical matching, we used BM25+ in package rank\_bm25 library \footnote{\url{https://pypi.org/project/rank-bm25/}}. For semantic searching, we trained sentence-transformer models by contrastive learning. For each query in training dataset, the label of the relevant article to query is 1 while the negative sample is 0. To get the negative samples, we took the top-k highest ranking score in the previous training round. Then the samples for each query was k+(number of positive articles)as positive(pos) and negative(neg) pairs. We used BM25+ and then trained the sentence-transformer model for three rounds. Our pipeline for training are shown in Figure \ref{pipeline for training} 

\section{Experiments}
\subsection{Datasets}
The dataset used in our paper is from the original dataset in the paper published by Kien et al. \cite{kien-etal-2020-answering}. 
We clean the data to reduce the noise in the legal corpus. The removed laws and articles do not appear in QA dataset while QA dataset remains the original so that data cleaning does not affect the evaluation results.
Finally, we obtained the legal corpus  containing 8,436 documents with 114,177 articles.
In legal corpus, we concatenated title and text in each article as a "very long" sentence. 

\begin{table}
\parbox{.45\linewidth}{
\centering
\caption{Distribution of articles length by syllables}
\label{len_syl}
\begin{tabular}{ccc}
        \hline
         Length & Amount & Proportion\\
        \hline
        < 100 & 32950 & 28.86\% \\
        \hline
        101 - 256 & 43923 & 38.47\% \\ 
        \hline
        257 - 512 & 23799& 20.84\%\\
        \hline
        513+ & 9483 & 11.83\% \\ 
        \hline
\end{tabular}

}
\hfill
\parbox{.45\linewidth}{
\centering
\caption{Distribution of articles length by words}
\label{len_word}
\begin{tabular}{ccc}
        \hline
         Length & Amount & Proportion\\
        \hline
        < 100 & 43763 & 38.33\% \\
        \hline
        101 - 256 & 42800 & 37.48\% \\ 
        \hline
        257 - 512 & 18636& 16.33\%\\
        \hline
        513+ & 6111 & 7.86\% \\ 
        \hline
\end{tabular}
}
\end{table}
Not like English, Vietnamese syllables and word tokens are different. For example, 4-syllable written text "Bài báo khoa học" (scientific paper) form 2 words "Bài\_báo \raisebox{-0.4ex}{\footnotesize{paper}}  khoa\_học \raisebox{-0.4ex}{\footnotesize{scientific}}". We used VNCoreNLP \cite{vu-etal-2018-vncorenlp} for word segmentation. We divided the corpus into 4 types of lengths: < 100, 101-256,257-512, and 513+. 256 is the maximum number of tokens supported by PhoBERT while 512 is the maximum number of tokens supported by ViBERT. Table \ref{len_syl} and table \ref{len_word} show percentage of articles based on length.

For QA dataset, we found that only 1,709 articles (about 1.5\% of the whole articles) in the legal corpus appear and about 95\% of questions  in both training and testing datasets have one to three relevant articles.
\subsection{Experimental procedure}

\begin{figure}[!h]
    \raggedright 
    \centering
    \includegraphics[scale=0.6]{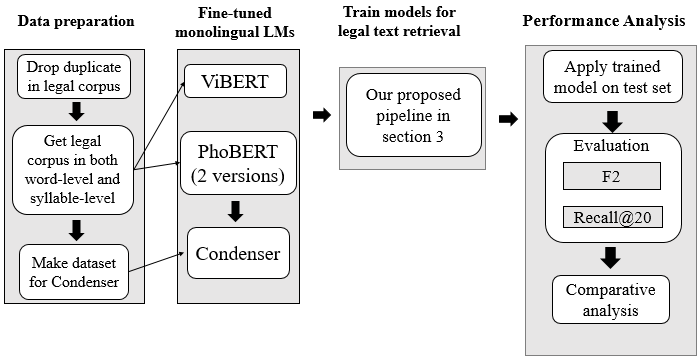}
    \caption{Experiment procedure}
    \label{experiment procedure}
\end{figure}

We use a single NVIDIA Tesla P100 GPU via Google Colaboratory to train all the models. Figure \ref{experiment procedure} presents an overview of the experimental procedure in this paper.

\subsubsection{Monolingual pre-trained models}

\begin{itemize}
  \item PhoBERT  pre-training approach is based on RoBERTa \cite{Liu2019RoBERTaAR} which is using Dynamic Masking to create masked tokens. Because two PhoBERT versions are trained on large-scale Vietnamese dataset, they perform great results on several NLP tasks in Vietnamese like NER, POS tagging, and DP.
  \item ViBERT  leverages checkpoint from mBERT \cite{devlin2018bert} and continues training on 10 GB of Vietnamese news data.  ViBERT pre-training approach based on BERT which has model architecture is a multilayer bidirectional Transformer encoder.  ViBERT now is supporting sequence lengths up to 512 tokens, this is an advantage compared to only 256 tokens supported by PhoBERT.
\end{itemize}

\subsubsection{Word-level and syllable-level approaches} In a field that contains many semantic challenges such as law, we want to evaluate how words and syllables of Vietnamese affect the results of models. We trained transformer models on both word-level and syllable-level independently  in every below step. 
\subsubsection{Fine-tuning monolingual language models} We used our legal corpus to fine-tune masked language modeling of monolingual language models. We set gradient\_accumulation\_steps = 4, train\_batch\_size = 8, eval\_batch\_size = 8 and epochs = 20.
\subsubsection{Fine-tuning Condenser} After training mask language modeling, we used both versions of PhoBERT to train Condenser. We used source code \footnote{\url{ https://github.com/luyug/Condenser}} for creating data in Condenser's form and fine-tuning Condenser from PhoBERT.
\subsubsection{Lexical matching} For lexical matching, we used BM25+. VNCoreNLP was used for word segmentation. Stopwords were also removed to make the model focus on important words. Our goal after this step is to get the list of the top highest lexical similarity articles for each query.

\subsubsection{Training sentence-transformer }  We trained 3 rounds for sentence-transformer. The choosing method we mentioned above, for short, we will call the number of negative sample each round is k. In the first round, we picked 35 negative samples from BM25+. These negative samples almost are high lexical similarity to queries but low semantic similarity. In the second round, we picked 20 negative samples from the first round. These negative samples have better semantic similarity to queries. In the third round, we pick 15 negative samples from the second round. These negative samples are really close semantic to queries. We set batch\_size=8, epochs = 4, learning\_rate = 1e-5. For loss function, we used contrastive loss \cite{hadsell2006dimensionality}

\subsubsection{Evaluation metrics}
The first measure we use to evaluate the performance of the system is (macro)recall@20, where 20 is the number of the top selected articles.
The second evaluation metric we use is F2. The F2-measure is shown below:

\[Precision\raisebox{-0.7ex}{\footnotesize{i-th}} =  \frac{\text{the number of correctly retrieved articles of query i-th}}{\text{the number of retrieved articles of query i-th}} \]
\[Recall\raisebox{-0.7ex}{\footnotesize{i-th}} =  \frac{\text{the number of correctly retrieved articles of query i-th}}{\text{the number of relevant articles of query i-th}} \]
\[F2\raisebox{-0.7ex}{\footnotesize{i-th}} =  \frac{\text{5*Precision\raisebox{-0.7ex}{\footnotesize{i-th}}*Recall\raisebox{-0.7ex}{\footnotesize{i-th}}}}{\text{4*Precision\raisebox{-0.7ex}{\footnotesize{i-th}}+Recall\raisebox{-0.7ex}{\footnotesize{i-th}}}} \]
\[F2=  \text{average of (F2\raisebox{-0.7ex}{\footnotesize{i-th}})} \]

\subsection{Results of BM25+ and sentence-transformer}
The models we selected can only compute scores between query-rule pairs. The performance of the model on F2 will decrease if we choose too many articles. In order to get good results on the F2 measure, we have to set a suitable threshold. Let the highest score between each query and all articles as highest\_score. The articles would be chosen if their scores in range [highest\_score-threshold, highest\_score]. For short, we rewrite sentence-transformer  Condenser trained from PhoBERT-base and PhoBERT-large respectively SConPBB and SConPBL in the below table.
\begin{table}
\caption{Results of BM25+ and sentence-transformer}
\label{single model}
        \centering
        \begin{tabular}{|l|l|l|l|c|l|c|} 
        \hline
         \textbf{Model}  & \textbf{Recall@20} & \textbf{F2} \\
        \hline
        BM25+ & 0.557 & 0.221 \\
        \hline
        SPhoBERT-base(syl) & 0.944      & 0.700\\ 
        \hline
        SConPBB(syl)  & 0.953      & 0.697       \\
        \hline
        SPhoBERT-large(syl) & 0.940      & 0.715\\ 
        \hline
        SConPBL(syl)  & 0.953       & 0.719      \\
        \hline
        SViBERT(syl) & 0.942      & 0.679\\ 
        \hline
        SPhoBERT-base(word) & 0.954      & 0.721\\ 
        \hline
        SConPBB(word)  & 0.957      & 0.704       \\
        \hline
        SPhoBERT-large(word) & 0.954      & 0.727\\ 
        \hline
        SConPBL(word)  & 0.954       & 0.724      \\
        \hline
        SViBERT(word) & 0.921      & 0.669\\ 
        \hline
        \end{tabular}
\end{table}

The result in Table \ref{single model} shows that: sentence-transformer models outperform BM25+. However, we want to build a retrieval system with both lexical matching and semantic searching approaches. We consider the combining score of these two approaches. More than 95\% of questions have one to three relevant articles. We visualized the average score of the top 3 highest scores of BM25+. 

\begin{figure}[!h]
    \centering
    \includegraphics[scale=0.7]{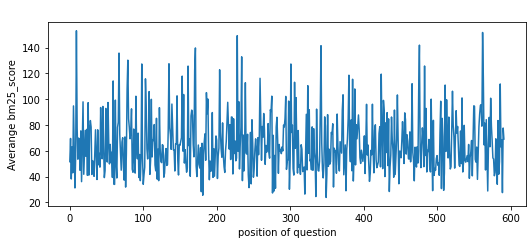}
    \caption{Average score of top 3 highest scores of BM25+}
    \label{fig:top3_BM25_avg}
\end{figure}

\subsection{Retrieval system using BM25+ and sentence-transformer}
Sentence-transformer uses the cosine similarity(cos\_sim) of the two vectors query and article to rank.Cos\_sim is always in range [-1,1]. According to Figure \ref{fig:top3_BM25_avg} average score of BM25+ is much larger than cos\_sim while semantic approach makes a big gap. Combining score for ranking in previous work like (1- $\alpha $)*lexical\_score + $\alpha$ * semantic\_score will not work or take much time to fine the $\alpha$. We proposed two combining score: sqrt(BM25\_score)*cos\_sim and BM25\_score*cos\_sim to rank our retrieval systems. Similar to single model, we used suitable threshold for retrieval system to get best results on F2. For short, we named the retrieval system by a name of sentence-transformer model within it.

\begin{table}[!h]
\caption{Results of retrieval systems }
\label{retrieval_sys}
        \centering
        \begin{tabular}{|l|c|c|c|c|c|c|}
        \hline
        &\multicolumn{2}{| c |}{sqrt(BM25\_score)*cos\_sim} &\multicolumn{2}{| c |}{BM25\_score*cos\_sim}\\
        \hline
         \textbf{System}    & \textbf{Recall@20} & \textbf{F2} & \textbf{Recall@20} & \textbf{F2} \\
        \hline
        SPhoBERT-base(syl) & 0.958 & 0.715 & 0.960 & 0.703\\ 
        \hline
        SConPBB(syl) & 0.967 & 0.712 & 0.965 & 0.692   \\
        \hline
        SPhoBERT-large(syl) & 0.951 & 0.726 & 0.955 & 0.706\\ 
        \hline
        SConPBL(syl) & 0.953 & 0.729 & 0.956 & 0.710   \\
        \hline
        SViBERT(syl) & 0.951 & 0.695 & 0.963 & 0.684\\ 
        \hline

        \hline
        SPhoBERT-base(word) & 0.963      & 0.725 & 0.964      & 0.706\\ 
        \hline
        SConPBB(word) & 0.962      & 0.719   & 0.964      & 0.700     \\
        \hline
        SPhoBERT-large(word) & 0.967      & \textbf{0.741} & \textbf{0.970}      & 0.721\\ 
        \hline
        SConPBL(word) & 0.960       & 0.738     & 0.967       & 0.718   \\
        \hline
        SViBERT(word) & 0.948      & 0.683 & 0.950      & 0.661\\ 
        \hline
        
        \end{tabular}
\end{table}

Based on the results from Tables \ref{single model}, \ref{retrieval_sys},  we had three observations:
\begin{itemize}
  \item If LMs were pre-trained on word-level data like PhoBERT and CondenserPhoBERT,  trained sentence-transformer models on word-level data bring better results. 
  \item If LM were pre-trained on syllable-level data like ViBERT, trained sentence-transformer model on syllable-level data bring better results.
  \item The system with both lexical matching and semantic searching performances better than sentence-transformer model within it.
  \item Sqrt(BM25\_score)*cos\_sim is the best ranking score for the system on F2 while BM25\_score*cos\_sim is the best on recall@20. 

\end{itemize}
\subsection{Comparison with previous Vietnamese legal text retrieval systems}
Our best performing system on F2 is SPhoBERT-large(word) system with ranking score: sqrt(BM25\_score)*cos\_sim. On Recall@20, our best performing system is SPhoBERT-large(word) system with ranking score: BM25\_score*cos\_sim. For short, we call our best retrieval system on F2 is our best system 1 and our best retrieval system on Recall@20 is our best system 2.
\begin{table}
\parbox{.45\linewidth}{
\centering
\caption{Comparison with previous Vietnamese legal text retrieval systems on F2}
        \label{CompareF2}
        \begin{tabular}{|l|c|} 
        \hline
        \textbf{System}  & \textbf{F2}  \\
        \hline
        Our best system 1 & \textbf{0.741}\\ 
        \hline
        Attentive CNN \cite{nguyen2022toward} & 0.4774  \\
        \hline
        XLM-RoBERTa \cite{nguyen2022toward} & 0.2006\\ 
        \hline
\end{tabular}

}
\hfill
\parbox{.45\linewidth}{
\centering
\caption{Comparison with previous Vietnamese legal text retrieval systems on Recall@20}
        \label{CompareRecall@20}
        \begin{tabular}{|p{3.5cm}|c|} 
        \hline
        \textbf{System} & \textbf{Recall@20}  \\
        \hline
        Our best system 2 & \textbf{0.970}\\ 
        \hline
        ElasticSearch + Attentive CNN \cite{kien-etal-2020-answering}  & 0.825  \\
        \hline
        Birch(256 first words) \cite{kien-etal-2020-answering} & 0.763\\
        \hline
        Birch(title) \cite{kien-etal-2020-answering} & 0.783\\
        \hline
\end{tabular}
}
\end{table}

Table \ref{CompareF2} and Table \ref{CompareRecall@20} illustrate the results of our best systems compared with the previous systems \cite{kien-etal-2020-answering,nguyen2022toward}. It is proved that our system yields the best performance on both metrics. 
\subsection{Analysis by the number of relevant articles related to queries}
We used best retrieval system from each sentence-transformer model to analysis effect of the number of relevant articles to queries in both metrics. 
\begin{figure}[!h]
    \raggedright 
    \centering
    \includegraphics[scale=0.45]{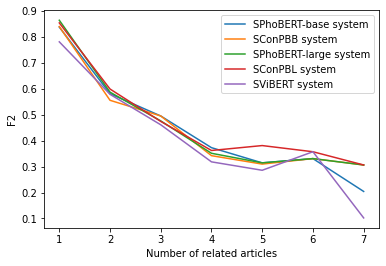}
    \caption{Effect of the number of relevant articles to queries on F2}
    \label{F2_num}
\end{figure}

\begin{figure}[!h]
    \raggedright 
    \centering
    \includegraphics[scale=0.45]{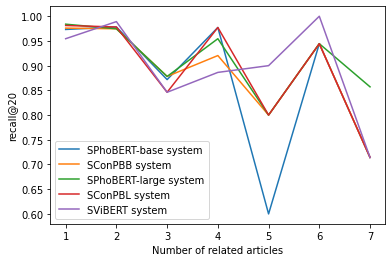}
    \caption{Effect of the number of relevant articles to queries on recall@20}
    \label{recall_num}
\end{figure}

Our observations based on the results from Figures \ref{F2_num}, \ref{recall_num}:
\begin{itemize}
  \item The more relevant articles relate to the query, the lower F2 of the systems there. Part of the reason for this is because we set the threshold pretty low. So the range [highest\_score-threshold, highest\_score] would not be large enough to cover many the relevant articles.
  \item Queries with (3,5,7) relevant articles make the systems with recall@20 significantly lower than queries with (2,4,6) relevant articles. Most of the recall@20 results of systems are pulled down because of queries with (2,4,6) relevant articles.
  \item SViBERT system achieves much better results than other systems on the recall@20 measure in queries with (5,6) relevant articles.
  \item SPhoBERT-large achieved the best results because it outperformed other systems in queries with (3,5,7) relevant articles. 

\end{itemize}

\subsection{Data-driven error analysis}
There are two main errors we found during the experiment.
First, wrong crawling relevant articles for query in the dataset. After a thorough discussion, we all agreed that the answer to this question was wrong. The example in Table \ref{tab:wrong_crawling} shows that relevant article was wrongly collected. The question is about person who is obliged to execute the judgment returns the non-implementation papers while relevant article is about principles of auction of land use rights. We did not find any correlation between the question and the answer.
\begin{table}
\caption{Example of wrong crawling relevant articles}
    \centering
        \begin{tabularx}{\textwidth}{|>{\hsize=.15\hsize\centering\arraybackslash}X|X|>{\hsize=.1\hsize\centering\arraybackslash}X|}
        \cline{1-2}
        \textbf Q/A/AC &  \multicolumn{1}{c|}{\textbf{Content}}\\
        \cline{1-2}
        Query(Q) & Người có nghĩa vụ phải thi hành án trả giấy tờ không thực hiện thì cơ quan thi hành án dân sự cưỡng chế như thế nào ? \textit({If the person who is obliged to execute the judgment returns the non-implementation papers, how does the civil judgment enforcement agency coerce ?)}\\
        \cline{1-2}
        Answer(A) & Article 117 in law id:45/2013/QH13 \\
        \cline{1-2}
        Article content (AC) & Nguyên tắc đấu giá quyền sử dụng đất \textit({Principles of auction of land use rights)} \newline 1 . Đấu giá quyền sử dụng đất được thực hiện công khai , liên tục , khách quan , trung thực , bình đẳng , bảo vệ quyền và lợi ích hợp pháp của các bên tham gia. \textit({Auction of land use rights is conducted publicly, continuously, objectively, honestly, equally, protecting the legitimate rights and interests of the participating parties)} \newline 2. Việc đấu giá quyền sử dụng đất phải đúng trình tự , thủ tục theo quy định của pháp luật về đất đai và pháp luật về đấu giá tài sản\textit({The auction of land use rights must comply with the order and procedures prescribed by the law on land and the law on asset auction.)} \\
        \cline{1-2}
        \end{tabularx}
    \label{tab:wrong_crawling}
\end{table}

Second, very hard semantic queries and relevant articles. These queries often miss-matching lexical with their relevant articles. Both of them are also long and contain difficult specialized words. To be able to understand the semantic of these queries and articles will still be a big challenge for retrieval systems.
\begin{table}[!h]
    \centering
    \caption{Example of very hard semantic query and relevant articles}
    \label{tab:hard_semantic}
        \begin{tabularx}{\textwidth}{|>{\hsize=.15\hsize\centering\arraybackslash}X|X|>{\hsize=.1\hsize\centering\arraybackslash}X|}
        \cline{1-2}
        \textbf Q/A/AC &  \multicolumn{1}{c|}{\textbf{Content}}\\
        \cline{1-2}
        Query(Q) & Có thể kê biên tài sản riêng của giám đốc để đảm bảo trả nợ cho công ty không ? \textit({Can the director's personal assets be distraint to ensure the repayment of the company's debt ?)}\\
        \cline{1-2}
        Answer(A) & Article 74 in law id:91/2015/QH13 \\
        \cline{1-2}
        Article content (AC) & Pháp nhân \newline 1.  Một tổ chức được công nhận là pháp nhân khi có đủ các điều kiện sau đây \textit{(An organization is recognized as a "pháp nhân" when the following conditions are satisfied)} a ) Được thành lập theo quy định của Bộ Luật này , luật khác có liên quan; \textit({a) Established under the provisions of this law and other relevant laws;)} b ) Có cơ cấu tổ chức theo quy định tại Điều 83 của bộ luật này;\textit{(b) Having an organizational structure as prescribed in Article 83 of this law;)} c) Có tài sản độc lập với cá nhân , pháp nhân khác và tự chịu trách nhiệm bằng tài sản của mình ; \textit{(c) Having assets independent of other individuals or legal entities and taking responsibility for their own property;)}   \textbf{c) Có tài sản độc lập với cá nhân , pháp nhân khác và tự chịu trách nhiệm bằng tài sản của mình ; \textit{(c) Having assets independent of other individuals or legal entities and taking responsibility for their own property;)}} d ) Nhân danh mình tham gia quan hệ pháp luật một cách độc lập. \textit{(d) Independently participate in legal relations in their own name.)} \newline 2 . Mọi cá nhân , pháp nhân đều có quyền thành lập pháp nhân , trừ trường hợp luật có quy định khác . \textit{(2 . Every individual and juridical person has the right to establish a juridical person, unless otherwise provided for by law.)}\\
        \cline{1-2}
        \end{tabularx}
    
\end{table}

In Table \ref{tab:hard_semantic}, "pháp nhân" is a specialized word in Vietnamese legal. To the best of our knowledge, the word has the closest meaning to it in English is corporation. We found the answer for the query in clause c) of this article. This article and query are lexical mismatching, so retrieval system will not work as well as sentence-transformer model. We used SPhoBERT-large which is the best semantic searching model to evalute cos\_sim between question and clause c) only. The result is 0.351. This shows that our model is still limited in the face of questions and answers with high semantic difficulty.
\section{Conclusion and future work}
In this paper, we proposed multi-stage information retrieval based on sentence-transformers for Vietnamese legal texts. We also compare the best performance model to the Attentive CNN model (previous SOTA in this QA dataset). The results indicate that our models outperform previous SOTA in both evaluation metrics.
 
In future work, we will try other techniques for long articles like summary, and keyword extraction. We will also improve the models to learn the to learn the relationship between queries and rules of very high semantic complexity.
%
%
%

\bibliographystyle{splncs04}
\bibliography{citiation}

\end{document}